\documentclass[sigconf,9pt]{acmart}

\usepackage{amsmath}

\usepackage{amssymb}
\usepackage{graphicx}
\usepackage{booktabs}
\setcopyright{acmlicensed}
\copyrightyear{2026}
\acmYear{2026}
\acmDOI{10.1145/XXXXXX.XXXXXX} 
\acmISBN{978-1-4503-XXXX-X/26/06}

\renewcommand\footnotetextcopyrightpermission[1]{}

\title{
Learning Discriminative Signed Distance Functions from Multi-scale Level-of-detail Features for 3D Anomaly Detection
}

\author{HaiBo Xiao}    
\affiliation{%
  \institution{College of Computer Science and
Software Engineering, Shenzhen
University}
  \city{Shenzhen}
  \country{China}}
\email{2400101026@email.szu.edu.cn}

\author{Hanzhe Liang}    
\affiliation{%
  \institution{Biological and Life Sciences Division, Mohamed bin Zayed University of Artificial Intelligence, UAE\\Shenzhen Audencia Financial Technology Institute, Shenzhen University, China}
  \city{Shenzhen}
  \country{China}}
\email{2023362051@email.szu.edu.cn}

\author{Jie Zhou}    
\affiliation{%
  \institution{School of Artificial Intelligence, Shenzhen University \\ Guangdong Provincial Key Laboratory of Intelligent Information Processing}
  \city{Shenzhen}
  \country{China}}
\email{davidgao@email.szu.edu.cn}

\author{Jinbao Wang}    
\affiliation{%
  \institution{School of Artificial Intelligence, Shenzhen University \\ Guangdong Provincial Key Laboratory of Intelligent Information Processing}
  \city{Shenzhen}
  \country{China}}
\email{wangjb@email.szu.edu.cn}

\author{Can Gao}    
\affiliation{%
  \institution{College of Computer Science and
Software Engineering, Shenzhen University \\
Guangdong Provincial Key Laboratory of Intelligent Information Processing}
  \city{Shenzhen}
  \country{China}}
\email{davidgao@email.szu.edu.cn}

\begin{abstract}
Detecting anomalies from 3D point clouds has received increasing attention in the field of computer vision, with some group-based or point-based methods achieving impressive results in recent years. However, learning accurate point-wise representations for 3D anomaly detection faces great challenges due to the large scale and sparsity of point clouds. In this study, a surface-based method is proposed for 3D anomaly detection, which learns a discriminative signed distance function using multi-scale level-of-detail features. We first present a Noisy Points Generation (NPG) module to generate different types of noise, thereby facilitating the learning of discriminative features by exposing abnormal points. Then, we introduce a Multi-scale Level-of-detail Feature (MLF) module to capture multi-scale information from a point cloud, which provides both fine-grained local and coarse-grained global feature information. Finally, we design an Implicit Surface Discrimination (ISD) module that leverages the extracted multi-scale features to learn an implicit surface representation of point clouds, which effectively trains a signed distance function to distinguish between abnormal and normal points. Experimental results demonstrate that the proposed method achieves an average object-level AUROC of 92.1\% and 85.9\% on the Anomaly-ShapeNet and Real3D-AD datasets, outperforming the current best approach by 2.1\% and 3.6\%, respectively. Codes are available at https://anonymous.4open.science/r/DLF-3AD-DA61.
\end{abstract}

\begin{document}
\maketitle

\section{Introduction}
3D anomaly detection (AD)~\cite{liang20253dadsurvey,  lin2025survey} aims to identify points and regions that deviate from normal distributions of point clouds and has recently gained widespread attention due to its potential in high-precision industrial inspection and autonomous driving~\cite{liu2025corenet,du2022tensor}. Learning from normal point clouds is a very appealing solution because it does not require collecting a large number of training samples and also avoids high-cost anomaly labeling\cite{3D-ST,liang2025lightweight,bhunia2025odd}. As a result, in 3D anomaly detection, exploring effective representation methods for normal point clouds has become a primary research topic because of its crucial role in determining detection performance~\cite{cohen2022semi}. 

\begin{figure}[tt]
\centering
\includegraphics[width=\columnwidth]{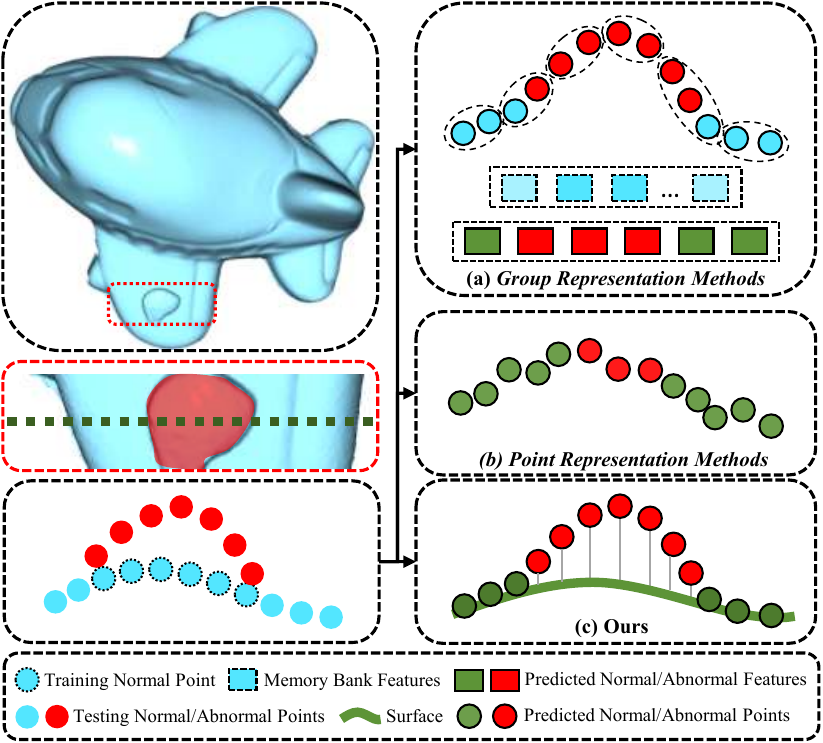}
\caption{
Comparison between previous methods and ours. (a) Group representation methods primarily rely on the extracted group-level normal features, which lack fine-grained details required for precise anomaly localization. (b) Point representation methods focus solely on individual point-level features while neglecting global geometric structures and local topological relationships. (c) Our method employs a surface representation that effectively captures global shape and local neighborhood information, enabling superior performance in both anomaly detection and localization. 
}
\label{solution}
\end{figure}

Existing approaches for 3D anomaly detection\cite{GS-CLIP} employ either group-based or point-based methods to represent normal point clouds. Group-based methods, such as BTF~\cite{BTF}, Reg3D-AD \cite{Real3D-AD}, and Group3AD~\cite{Group3AD}, separate each point cloud into several groups and learn a group-level representation. Anomaly detection is then performed based on the reconstruction error or feature comparison of each group. In contrast, point-based methods, such as IMRNet~\cite{IMRNet}, PO3AD~\cite{PO3AD}, and DUS-Net~\cite{DU-Net}, try to predict the coordinate or offset of each point and identify points with significant errors as anomalies.

However, these feature representations may have their limitations. Group feature representation lacks detailed features for accurate anomaly localization, while point feature representation ignores global geometric structure and local topological relationship information, leading to unsatisfactory performance. Intuitively, an effective 3D anomaly detection framework requires a unified representation that simultaneously captures global and local geometric structure information. Surface-based representation, such as Signed Distance Function (SDF)~\cite{park2019deepsdf}, emerges as a compelling solution to this challenge. Unlike separated groups or isolated points, modeling the continuous surface inherently captures both high-fidelity local topology and complete global geometry of point clouds. To this end, we propose a surface-based method to learn a discriminative SDF by exploiting the multi-scale level-of-detail features of point clouds. Figure \ref{solution} illustrates the differences between existing methods and ours. To sum up, the main contributions of this study are as follows:

\begin{itemize}

\item We propose a Noisy Points Generation (NPG) module to generate surface and noisy points, thereby facilitating the extraction of discriminative features.

\item We introduce a Multi-scale Level-of-detail Feature (MLF) module to extract multi-scale features, providing rich local and global information for representation.

\item We present an Implicit Surface Discrimination (ISD) module to learn an accurate surface representation for point clouds, which trains a discriminative signed distance function using the extracted multi-scale features to identify anomalous points.

\item Extensive comparative experiments on Anomaly-ShapeNet and Real3D-AD demonstrate the superiority of our method, achieving an average object-level AUROC of 92.1\% and 85.9\%, and an average point-level AUROC of 92.4\% and 85.2\%, respectively.

\end{itemize}
\section{Related Work}
\label{relatedwork}
The primary goal of 3D anomaly detection~\cite{chen2021rapid,tao2023anomaly,Rani_2024} is to identify and localize anomalies within point cloud data. Existing approaches to this task can be broadly categorized into group-based and point-based methods. 
\subsection{Group-based Methods }
Group-based method aims to extract group-level features and identify anomalies by comparing with the stored memory features or reconstructed features. Reg3D-AD \cite{Real3D-AD} employed a dual-feature representation to construct a memory bank of normal prototypes via coreset sampling, and anomaly scores were calculated by comparing the extracted features of the registered test point clouds and the memory bank. Group3AD \cite{Group3AD} improved anomaly detection performance by introducing the intercluster uniformity and intracluster alignment networks to enhance the group-level feature separation between clusters and tighten the distribution within clusters. 
M3DM \cite{M3DM} aligned features extracted from RGB and 3D point clouds and constructed multi-type feature memory banks for multi-modal anomaly detection and localization. 
By generating 2D modalities from 3D point clouds, Looking3D\cite{Bhunia2024look}, CPMF \cite{CPMF}, and ISMP \cite{ISMP} capitalized on additional information to enhance the ability of anomaly detection.

\subsection{Point-based Methods }
Point-based methods extract point-level features to construct models for predicting or reconstructing point coordinates, and the coordinate offsets are considered the point anomaly scores. 
IMRNet \cite{IMRNet} learned the normal local structures of point clouds through masked coordinate prediction and detected anomalies by leveraging the discrepancies between input and reconstructed coordinates. R3D-AD \cite{R3DAD} employed PointMAE \cite{pointmae} and RANSAC \cite{ransac} to extract registration features and localized anomalies via noise diffusion. PO3AD \cite{PO3AD} trained a predictor using normal vector-guided pseudo-anomalies and identified anomalies through the predicted offsets. DUS-Net \cite{DU-Net} employed an encoder-decoder architecture to reconstruct point clouds by preserving the geometric structure of group centers and detected anomalies by comparing features between up-sampled points from group centers and the raw points.

In addition, several approaches have explored 3D anomaly detection from the perspectives of multi-modal~\cite{CPMF}, multi-view~\cite{splat, splat++}, multi-categories~\cite{mc3d,lu2025c3dadcontinual3danomaly, Simple3D}, large language models \cite{pointad,wang20253dzal,tang2025exploringpotentialencoderfreearchitectures,xu2025zeroshotanomalydetectionreasoning}, and zero-shot~\cite{GS-CLIP}, achieving encouraging results. However, both group-based and point-based methods struggle to effectively integrate global and local features of point clouds, which hinders the models from obtaining high performance. Although PASDF ~\cite{PASDF} attempted to learn the surface information of 3D point clouds for anomaly detection, its performance is still limited by only using normal points and single-scale features. To address these problems, this study proposes a novel surface-based method that leverages multi-scale features to train a discriminative SDF network for accurate anomaly detection.

\section{Approach}
\begin {figure*}[tt]
\centering
\includegraphics [width=\linewidth]{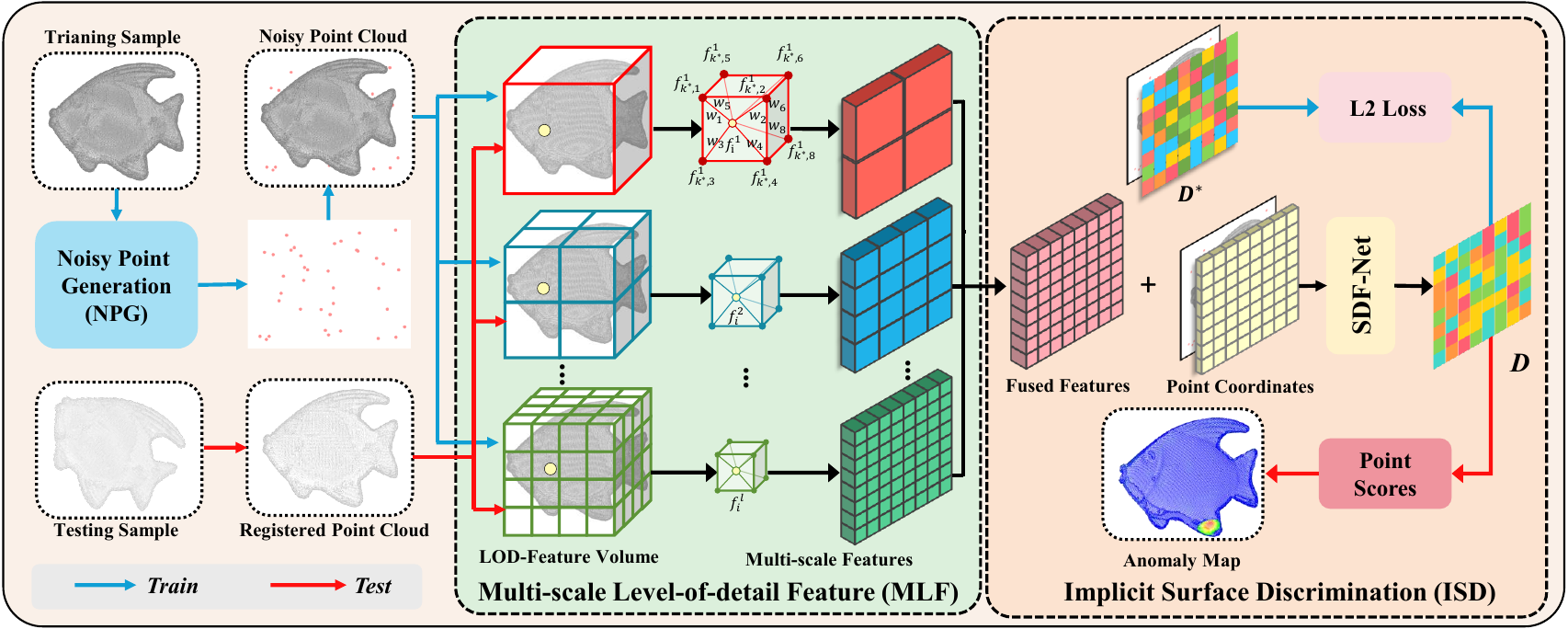}
\caption {The overall framework of our method. The Noisy Point Generation (NPG) module is first used to generate surface and noisy points from input point clouds. Then, the Multi-scale Level-of-detail Feature (MLF) module is employed to extract multi-scale features from the raw point cloud with generated points. Finally, the Implicit Surface Discrimination (ISD) module is designed to fuse these multi-scale features and combine them with point coordinates to train an SDF-Net. The network is optimized with L2 loss between the signed distance values predicted by SDF-Net and the ground-truth values. In inference, the predicted signed distance values of the testing point cloud are regarded as point-level anomaly scores.}
\label {DLF3AD_pipeline}
\end {figure*}

To address the limitations of feature representation in existing approaches and enhance the capability to discriminate between normal and abnormal points, we propose a surface-based method for 3D anomaly detection. As illustrated in Figure~\ref{DLF3AD_pipeline}, the proposed method consists of three core modules: Noisy Points Generation (NPG), Multi-scale Level-of-detail Feature (MLF), and an Implicit Surface Discrimination (ISD) module. Details of each module are elaborated in the following sections.

\subsection {Noisy Point Generation}
In 3D anomaly detection, models are typically trained on a limited set of normal point clouds. This poses a significant challenge in learning a distinct decision boundary between normal and abnormal points. To overcome this limitation, we propose a Noisy Point Generation (NPG) module, which synthesizes noisy points to facilitate the learning of more discriminative features for accurate anomaly detection.

Specifically, given a point cloud with $n$ points $ P = \{p_1, p_2, \cdots, p_n\} \in \mathbb{R}^{n \times 3}$ and its corresponding triangular mesh $\mathbf{\Delta}$, the point cloud is first normalized into the space $ \mathbb{B} = [-1,1]^3 $ to achieve uniform spatial scaling. The triangular mesh effectively captures the local structure of the point cloud and can serve as the core basis for noisy point generation. For each triangular mesh $\Delta_i \in \mathbf{\Delta}$, the sampling probability on this mesh is proportional to its area:
\begin{equation}
Pr(\Delta_i) = \frac{Area(\Delta_i)}{\sum_{\Delta_j \in \mathbf{\Delta}} Area(\Delta_j)} ,
\end{equation}
where $Area$ means the area of a triangular mesh.

To overcome the sparseness and non-uniformity of raw point clouds, surface points are generated on the triangular mesh to form uniformly distributed normal points. For each triangular face defined by vertices $\Delta_i=\{p_1, p_2, p_3\}$, two random numbers $u_{1}, u_{2} \sim U([0,1])$ following a uniform distribution are determined, and each surface point is computed by: 
\begin{equation}
p_{\text{surf}} = (1 - \mu - \nu) \cdot p_1 + \mu \cdot p_2 + \nu \cdot p_3,
\end{equation}
where $\mu=1-\sqrt{u_{1}}$ and $\nu=\sqrt{u_{1}}(1-\sqrt{u_{2}})$.

Surface points are generated based on the probabilities of triangular faces. These points inherently capture the underlying surface geometry of point clouds, ensuring a well-distributed set of normal points for training.

To promote the learning of discriminative features, two types of noisy points, namely near-surface noisy points and uniform noisy points, are further generated. Near-surface noisy points simulate small anomalies that surround the surface of point clouds, which are generated by adding Gaussian noise to surface points: 
\begin{equation}
p_{\text{near}} = p_{\text{surf}} + \epsilon , 
\end{equation}
where $\epsilon \sim \mathcal{N}(0, \sigma^2 \cdot \mathbf{I}) $ denotes 3D Gaussian noise, and $ \mathbf{I}$ means the identity matrix.

Additionally, uniform noisy points are generated to simulate large anomalies lying in the region far from the surface of point clouds, which are sampled uniformly within the normalized space $\mathbb{B}= [-1,1]^3$:
\begin{equation}
p_{\text{uni}} \sim {Sampling}_{\text{uni}}(\mathbb{B}), 
\end{equation}
where $Sampling_{\text{uni}}$ means the operator for uniform sampling from the space $\mathbb{B}$.

Finally, the normal surface points and the two types of noisy points are combined to form a set of points for training: 
\begin{equation}
\widetilde{P} = P_{\text{surf}} \cup P_{\text{near}} \cup P_{\text{uni}},
\end{equation}
where $P_{\text{surf}}$, $P_{\text{near}}$, and $P_{\text{uni}}$ denote the sets of surface, near-surface, and uniform points, respectively.

The sampling ratio among these types of points is controlled by a set of parameters $\mathbf{\alpha} = (\alpha_\text{surf}, \alpha_\text{near}, \alpha_\text{uni})$, whose effects will be analyzed in the experimental section. Meanwhile, each point $p_i$ in the point set $\widetilde{P}$ is assigned a corresponding ground-truth signed distance $d_i^{*}$, where the value of each surface point is set to 0, and the value of each noisy point is set to its directed distance to the surface. 

Empowered by the generated point set and its corresponding signed distance information, the model can learn an accurate SDF, thereby effectively distinguishing normal points from anomalies and achieving high anomaly detection precision.

\subsection{ Multi-scale Level-of-detail Feature }
Conventional 3D anomaly detection methods~\cite{pointad,PO3AD} typically rely on features extracted from a single scale. Nevertheless, these approaches may yield unsatisfactory results due to scale variations in the number of points and the size of different anomalies across datasets. Multi-scale information~\cite{tsai2022multi, Patchcore} has demonstrated significant benefits for various computer vision tasks. Extracting multi-scale features for 3D anomaly detection is a promising solution. To this end, we propose the Multi-scale Level-of-detail Feature (MLF) module, which effectively extracts and integrates both global geometric context and local structural details. 

Specifically, to obtain features from a generated point cloud $\widetilde{P}$ at different scales, $L$ levels of learnable 3D feature volumes $\{{V}^1, {V}^2, \dots$, ${V}^L\}$ are constructed. Low-level feature volumes, with coarser resolutions, are used to capture global features such as shape and topological structure. While higher-level volumes, with finer resolutions, accurately represent local details such as depressions and protrusions. These feature volumes at different levels exhibit scale complementarity.

For any feature volume ${V}^l$ at the $l$-th level, the normalized space $\mathbb{B}= [-1,1]^3$ is partitioned uniformly into a grid of voxels. The resolution of this grid is determined by $s_l = 2^{l + base\_lod}$, where $base\_lod$ is the base number of divisions along each axis. 
Each resulting voxel is defined by its 8 surrounding vertices and their corresponding learnable features, formally denoted as $V_k^l=(\textbf{v}_k^l, {f}_k^l)$, where $1\le k \le (s_l+1)^3$ and the dimension of each learnable feature is set to $32$. 

To mitigate the position error caused by noise and ensure the continuity and smoothness of features, trilinear interpolation~\cite{nielson} is used to compute the features of each point in the generated point cloud. 
Let $\widetilde{p}_i \in \widetilde{P}$ be a point in the generated point cloud, and $V_{k^*}^l = (\textbf{v}_{k^*}^l, \textbf{f}_{k^*}^l)$ be the voxel containing the point $\widetilde{p}_i$ at the $l$-th level. To obtain the point feature $f_i^l$, the Euclidean distances of the point $\widetilde{p}_i$ to each vertex of the voxel $V_{k^*}^l$ is first calculated, and the inverse distance is considered as the weight coefficient, which is defined as:
\begin{equation}
w_j = \left(1 - \frac{|x_i - x_j|}{d_l}\right) \left(1 - \frac{|y_i - y_j|}{d_l}\right) \left(1 - \frac{|z_i - z_j|}{d_l}\right),
\end{equation}
where $w_j$ means the weight of the the $j$-th vertex $\textbf{v}_{k^*, j}^l$ at the $l$-th level, $(x_i, y_i, z_i)$ and $(x_j, y_j, z_j)$ are the coordinates of the point $\widetilde{p}_i$ and the $j$-th vertex $\textbf{v}_{k^*, j}^l$, respectively, $d_l$ is the width of each voxel at the $l$-th level, and $|\cdot|$ represents the absolute value.
Then, the point feature $f_i^l$ is computed by a weighted fusion of the learnable features from the eight vertices:
\begin{equation}
f_i^l = \sum_{j=1}^{8} w_j \cdot f_{k^*, j}^l,
\end{equation}
where $f_{k^*, j}^l$ and $w_j$ denote the learnable features of the $j$-th vertex at the $l$-th level and its corresponding weight.

After calculating the features of all points, the features set $F^l$ at the $l$-th level can be obtained. By repeating the above process across all levels, the multi-scale features $F =\{F^1, F^2, \cdots, F^L\}$ are finally generated, which contain rich and hierarchical information that benefits the training of the anomaly detection model.

\subsection{ Implicit Surface Discrimination }
The extracted multi-scale features from the MLF module offer a comprehensive representation of the point cloud, capturing both global structure and local detail. The core challenge for anomaly detection is to leverage extracted features to distinguish between normal and abnormal points effectively. To this end, we design an Implicit Surface Discrimination (ISD) module to learn a surface representation of the point cloud, which fully exploits the extracted multi-scale features to train a Signed Distance Function (SDF) network for abnormal point identification.

Specifically, to fully leverage the multi-scale features, the features extracted by the MLF module across all levels are first fused into a powerful feature representation. This can be formalized as: 
\begin{equation}
\widetilde{F} = F^1 \oplus F^2 \oplus \cdots \oplus F^L,
\end{equation}
where $F^l$ is the features extracted from the $l$-th level, $\oplus$ denotes the element-wise addition operator, and $\widetilde{F}$ is the resulting fused feature representation for all points.

Subsequently, these fused features $\widetilde{F}$, along with the point's coordinate information $XYZ$, are fed into a discriminative SDF network (SDF-Net) to predict the signed distance value for each point. The SDF-Net is composed of a 4-layer perceptron network. The prediction process can be formulated as follows:
\begin{equation}
\widetilde{D} = SDF(Concat(\widetilde{F}, XYZ)),
\end{equation}
where $SDF$ denotes the learned SDF-Net, $Concat$ means the feature concatenation operator, and $\widetilde{D}$ represents the set of predicted signed distance values for all input points $\widetilde{P}$.

To optimize the SDF-Net, the Mean Squared Error (MSE) loss is introduced to quantify the difference between the predicted values and the ground truth values:
\begin{equation}
\label{eq:sdf_loss}
\mathcal{L}_{\text{SDF}} = \frac{1}{Card(\widetilde{P})} \sum_{\widetilde{p}_i \in \widetilde{P}} (\widetilde{d}_i - d^{*}_{i} )^2,
\end{equation}
where $\widetilde{d}_i$ and $d^{*}_{i}$ denote the signed distance value predicted by the SDF-Net and the ground-truth value for the point $\widetilde{p}_i$, respectively, and $Card(\cdot)$ represents the cardinality of a set.

By minimizing this loss, the network is forced to accurately represent the underlying normal surface. For abnormal points, they have large ground-truth distance values, and the network is also required to predict large values to minimize errors. Thus, the final SDF network is inherently trained to assign high signed distance values to abnormal points.

\subsection{Training and Inference}
\textbf{Training}. In the training phase, the NPG module is first used to generate different types of points and their ground-truth SDF values $d^{*}$. Subsequently, the MLF module extracts multi-scale features from the generated points. Finally, these features are fused and fed into the ISD module to train the discriminative SDF-Net. The entire model is optimized end-to-end by minimizing the SDF loss $\mathcal{L}_{SDF}$ (defined in Eq.~\ref{eq:sdf_loss}), which forces the network to accurately predict the signed distance value for each point.
\begin{table*}[!ht]
  \centering
  \resizebox{0.98\textwidth}{!}{
    \begin{tabular}{c|cccccccccccccc}
    \toprule
    \textbf{Method} & \textbf{ashtray0} & \textbf{bag0} & \textbf{bottle0} & \textbf{bottle1} & \textbf{bottle3} & \textbf{bowl0} & \textbf{bowl1} & \textbf{bowl2} & \textbf{bowl3} & \textbf{bowl4} & \textbf{bowl5} & \textbf{bucket0} & \textbf{bucket1} & \textbf{cap0} \\
    \midrule
\textbf{BTF (Raw) (CVPR23’)} & 0.578 & 0.410 & 0.597 & 0.510 & 0.568 & 0.564 & 0.264 & 0.525 & 0.385 & 0.664 & 0.417 & 0.617 & 0.321 & 0.564 \\
\textbf{BTF (FPFH) (CVPR23’)} & 0.420 & 0.546 & 0.344 & 0.546 & 0.322 & 0.509 & 0.668 & 0.510 & 0.490 & 0.609 & 0.699 & 0.401 & 0.633 & 0.618 \\
\textbf{M3DM (CVPR23’)} & 0.577 & 0.537 & 0.574 & 0.637 & 0.541 & 0.634 & 0.663 & 0.684 & 0.617 & 0.464 & 0.409 & 0.309 & 0.501 & 0.557 \\
\textbf{PatchCore (FPFH) (CVPR22’)} & 0.587 & 0.571 & 0.604 & 0.667 & 0.572 & 0.504 & 0.639 & 0.615 & 0.537 & 0.494 & 0.558 & 0.469 & 0.551 & 0.580 \\
\textbf{PatchCore (PointMAE)  (CVPR22')} & 0.591 & 0.601 & 0.513 & 0.601 & 0.650 & 0.523 & 0.629 & 0.458 & 0.579 & 0.501 & 0.593 & 0.593 & 0.561 & 0.589 \\
\textbf{CPMF (PR24’)} & 0.353 & 0.643 & 0.520 & 0.482 & 0.405 & 0.783 & 0.639 & 0.625 & 0.658 & 0.683 & 0.685 & 0.482 & 0.601 & 0.601 \\
\textbf{Reg3D-AD (NeurIPS23’)} & 0.597 & 0.706 & 0.486 & 0.695 & 0.525 & 0.671 & 0.525 & 0.490 & 0.348 & 0.663 & 0.593 & 0.610 & 0.752 & 0.693 \\
\textbf{IMRNet (CVPR24’)} & 0.671 & 0.660 & 0.552 & 0.700 & 0.640 & 0.681 & 0.702 & 0.685 & 0.599 & 0.676 & 0.710 & 0.580 & 0.771 & 0.737 \\
\textbf{R3D-AD (ECCV24’)} & 0.833 & 0.720 & 0.733 & 0.737 & 0.781 & 0.819 & 0.778 & 0.741 & 0.767 & 0.744 & 0.656 & 0.683 & 0.756 & 0.822 \\
\textbf{PO3AD (CVPR25')} & \textcolor{red}{1.000} & {0.833} & {\textcolor{blue}{0.900}} & {0.933} & {0.926} & {0.922} & 0.829 & 0.833 & 0.881 & \textcolor{red}{0.981} & 0.849 & \textcolor{blue}{0.853} & 0.787 & 0.877 \\
\textbf{DUS-Net (MM25')} & 0.867 & 0.605 & 0.838 & 0.871 & 0.827 & 0.844 & 0.869 & 0.952 & 0.839 & 0.859 & 0.776 & 0.838 & {\textcolor{blue}{0.822}} & 0.859 \\
\textbf{PASDF (ICCV25')} & \textcolor{red}{1.000} & \textcolor{blue}{0.995} & \textcolor{red}{1.000} & \textcolor{red}{1.000} & \textcolor{red}{1.000} & \textcolor{red}{1.000} & 0.948 & \textcolor{red}{1.000} & \textcolor{red}{1.000} & 0.933 & 0.912 & \textcolor{red}{0.968} & 0.775 & 0.852 \\
\textbf{CASL (AAAI26’)} & \textcolor{blue}{0.948} & 0.821 & 0.861 & \textcolor{blue}{0.990} & \textcolor{blue}{0.944} & \textcolor{blue}{0.930} & \textcolor{blue}{0.967} & \textcolor{blue}{0.995} & \textcolor{blue}{0.933} & 0.772 & \textcolor{blue}{0.981} & 0.821 & 0.783 & \textcolor{red}{0.914} \\
\textbf{Ours} & \textcolor{red}{1.000} & \textcolor{red}{1.000} & \textcolor{red}{1.000} & \textcolor{red}{1.000} & \textcolor{red}{1.000} & \textcolor{red}{1.000} & \textcolor{red}{0.970} & \textcolor{red}{1.000} & \textcolor{red}{1.000} & \textcolor{blue}{0.967} & \textcolor{red}{0.989} & \textcolor{red}{0.968} & \textcolor{red}{0.851} & \textcolor{blue}{0.896} \\


    \midrule
    \multicolumn{1}{c}{} &       &       &       &       &       &       &       &       &       &       &       &       &       &  \\
    \midrule
    \textbf{Method} & \textbf{cap3} & \textbf{cap4} & \textbf{cap5} & \textbf{cup0} & \textbf{cup1} & \textbf{eraser0} & \textbf{headset0} & \textbf{headset1} & \textbf{helmet0} & \textbf{helmet1} & \textbf{helmet2} & \textbf{helmet3} & \textbf{jar0} & \textbf{micro.} \\
    \midrule
\textbf{BTF (Raw) (CVPR23’)} & 0.527 & 0.468 & 0.373 & 0.403 & 0.521 & 0.525 & 0.378 & 0.515 & 0.553 & 0.349 & 0.602 & 0.526 & 0.420 & 0.563 \\
\textbf{BTF (FPFH) (CVPR23’)} & 0.522 & 0.520 & 0.586 & 0.586 & 0.610 & 0.719 & 0.520 & 0.490 & 0.571 & 0.719 & 0.542 & 0.444 & 0.424 & 0.671 \\
\textbf{M3DM (CVPR23’)} & 0.423 & 0.777 & 0.639 & 0.539 & 0.556 & 0.627 & 0.577 & 0.617 & 0.526 & 0.427 & 0.623 & 0.374 & 0.441 & 0.357 \\
\textbf{PatchCore (FPFH) (CVPR22’)} & 0.453 & 0.757 & {0.790} & 0.600 & 0.586 & 0.657 & 0.583 & 0.637 & 0.546 & 0.484 & 0.425 & 0.404 & 0.472 & 0.388 \\
\textbf{PatchCore (PointMAE)  (CVPR22')} & 0.476 & 0.727 & 0.538 & 0.610 & 0.556 & 0.677 & 0.591 & 0.627 & 0.556 & 0.552 & 0.447 & 0.424 & 0.483 & 0.488 \\
\textbf{CPMF (PR24’)} & 0.551 & 0.553 & 0.697 & 0.497 & 0.499 & 0.689 & 0.643 & 0.458 & 0.555 & 0.589 & 0.462 & 0.520 & 0.610 & 0.509 \\
\textbf{Reg3D-AD (NeurIPS23’)} & 0.725 & 0.643 & 0.467 & 0.510 & 0.538 & 0.343 & 0.537 & 0.610 & 0.600 & 0.381 & 0.614 & 0.367 & 0.592 & 0.414 \\
\textbf{IMRNet (CVPR24’)} & 0.775 & 0.652 & 0.652 & 0.643 & 0.757 & 0.548 & 0.720 & 0.676 & 0.597 & 0.600 & 0.641 & 0.573 & 0.780 & 0.755 \\
\textbf{R3D-AD (ECCV24’)} & 0.730 & 0.681 & 0.670 & 0.776 & 0.757 & 0.890 & 0.738 & 0.795 & 0.757 & 0.720 & 0.633 & 0.707 & 0.838 & 0.762 \\
\textbf{PO3AD (CVPR25')} & \textcolor{blue}{0.859} & \textcolor{blue}{0.792} & 0.670 & 0.871 & {0.833} & {\textcolor{blue}{0.995}} & {\textcolor{blue}{0.808}} & 0.923 & {0.762} & {\textcolor{blue}{0.961}} & \textcolor{blue}{0.869} & 0.754 & {0.866} & 0.776 \\
\textbf{DUS-Net (MM25')} & 0.775 & 0.736 & 0.739 & 0.869 & 0.776 & 0.644 & 0.741 & \textcolor{blue}{0.961} & 0.743 & 0.859 & {0.841} & {\textcolor{blue}{0.859}} & 0.744 & {\textcolor{blue}{0.837}} \\
\textbf{PASDF (ICCV25')} & 0.649 & 0.646 & 0.853 & \textcolor{blue}{0.971} & \textcolor{blue}{0.857} & 0.952 & \textcolor{red}{1.000} & 0.795 & 0.812 & 0.938 & 0.765 & 0.846 & \textcolor{red}{1.000} & \textcolor{red}{1.000} \\
\textbf{CASL (AAAI26’)} & \textcolor{red}{1.000} & \textcolor{red}{0.838} & \textcolor{red}{0.957} & 0.563 & \textcolor{red}{1.000} & 0.990 & 0.714 & 0.890 & \textcolor{blue}{0.818} & 0.943 & \textcolor{red}{0.954} & 0.761 & \textcolor{blue}{0.981} & 0.648 \\
\textbf{Ours} & 0.688 & 0.712 & \textcolor{blue}{0.919} & \textcolor{red}{1.000} & 0.538 & \textcolor{red}{1.000} & \textcolor{red}{1.000} & \textcolor{red}{1.000} & \textcolor{red}{0.849} & \textcolor{red}{0.990} & 0.713 & \textcolor{red}{0.945} & \textcolor{red}{1.000} & \textcolor{red}{1.000} \\

    \midrule
    \multicolumn{1}{c}{} &       &       &       &       &       &       &       &       &       &       &       &       &       &  \\
    \midrule
    \textbf{Method} & \textbf{shelf0} & \textbf{tap0} & \textbf{tap1} & \textbf{vase0} & \textbf{vase1} & \textbf{vase2} & \textbf{vase3} & \textbf{vase4} & \textbf{vase5} & \textbf{vase7} & \textbf{vase8} & \textbf{vase9} & \textbf{Average} \\
    \midrule
\textbf{BTF (Raw) (CVPR23’)} & 0.164 & 0.525 & 0.573 & 0.531 & 0.549 & 0.410 & 0.717 & 0.425 & 0.585 & 0.448 & 0.424 & 0.564 & 0.493 \\
\textbf{BTF (FPFH) (CVPR23’)} & 0.609 & 0.560 & 0.546 & 0.342 & 0.219 & 0.546 & 0.699 & 0.510 & 0.409 & 0.518 & 0.668 & 0.268 & 0.528 \\
\textbf{M3DM (CVPR23’)} & 0.564 & {0.754} & 0.739 & 0.423 & 0.427 & 0.737 & 0.439 & 0.476 & 0.317 & 0.657 & 0.663 & 0.663 & 0.552 \\
\textbf{PatchCore (FPFH) (CVPR22’)} & 0.494 & 0.753 & 0.766 & 0.455 & 0.423 & 0.721 & 0.449 & 0.506 & 0.417 & 0.693 & 0.662 & 0.660 & 0.568 \\
\textbf{PatchCore (PointMAE)  (CVPR22')} & 0.523 & 0.458 & 0.538 & 0.447 & 0.552 & 0.741 & 0.460 & 0.516 & 0.579 & 0.650 & 0.663 & 0.629 & 0.562 \\
\textbf{CPMF (PR24’)} & 0.685 & 0.359 & 0.697 & 0.451 & 0.345 & 0.582 & 0.582 & 0.514 & 0.618 & 0.397 & 0.529 & 0.609 & 0.559 \\
\textbf{Reg3D-AD (NeurIPS23’)} & 0.688 & 0.676 & 0.641 & 0.533 & 0.702 & 0.605 & 0.650 & 0.500 & 0.520 & 0.462 & 0.620 & 0.594 & 0.572 \\
\textbf{IMRNet (CVPR24’)} & 0.603 & 0.676 & 0.696 & 0.533 & 0.757 & 0.614 & 0.700 & 0.524 & 0.676 & 0.635 & 0.630 & 0.594 & 0.661 \\
\textbf{R3D-AD (ECCV24’)} & 0.696 & 0.736 & \textcolor{blue}{0.900} & 0.788 & 0.729 & 0.752 & 0.742 & 0.630 & 0.757 & 0.771 & 0.721 & 0.718 & 0.749 \\
\textbf{PO3AD (CVPR25')} & 0.573 & 0.745 & 0.681 & {0.858} & 0.742 & {0.952} & {0.821} & 0.675 & {\textcolor{blue}{0.852}} & \textcolor{blue}{0.966} & 0.739 & {0.830} & {0.839} \\
\textbf{DUS-Net (MM25')} & 0.688 & 0.739 & 0.840 & 0.833 & 0.808 & 0.644 & 0.766 & {0.749} & 0.838 & 0.844 & {0.808} & 0.525 & 0.797 \\
\textbf{PASDF (ICCV25')} & \textcolor{blue}{0.713} & \textcolor{red}{0.882} & 0.793 & \textcolor{red}{1.000} & \textcolor{blue}{0.929} & \textcolor{red}{1.000} & 0.806 & \textcolor{blue}{0.912} & \textcolor{red}{1.000} & \textcolor{red}{1.000} & \textcolor{blue}{0.924} & 0.836 & \textcolor{blue}{0.900} \\
\textbf{CASL (AAAI26’)} & \textcolor{red}{1.000} & \textcolor{blue}{0.881} & 0.554 & 0.832 & 0.888 & \textcolor{blue}{0.996} & \textcolor{red}{0.890} & \textcolor{red}{0.933} & 0.836 & 0.948 & \textcolor{red}{0.993} & \textcolor{red}{0.995} & 0.887 \\
\textbf{Ours} & 0.690 & 0.848 & \textcolor{red}{0.930} & \textcolor{blue}{0.933} & \textcolor{red}{0.967} & \textcolor{red}{1.000} & \textcolor{blue}{0.861} & 0.824 & \textcolor{red}{1.000} & \textcolor{red}{1.000} & \textcolor{blue}{0.924} & \textcolor{blue}{0.873} & \textcolor{red}{0.921} \\


    \bottomrule

    \end{tabular}%
}
    \caption{The O-AUROC ($\uparrow$) performance of different methods on Anomaly-ShapeNet, where the best and second-place results are highlighted in \textcolor{red}{red} and \textcolor{blue}{blue}, respectively.}
  \label{results1}
\end{table*}

\textbf{Inference.} 
During inference, each test point cloud is first normalized and registered to align with the training data. It is then processed by the MLF module to extract multi-scale features. These features after fusion are fed into the trained SDF-Net to predict a signed distance value $\widetilde{d}$ for each point. The absolute value of $\widetilde{d}$, i.e., $|\widetilde{d}|$, directly serves as the anomaly score, indicating the point's deviation from the learned normal surface. A larger magnitude suggests a higher likelihood of being an anomaly. Consequently, the point-level anomaly score is derived directly from the absolute predicted distance of each point, while the object-level score is defined as the maximum score across all points in the point cloud.
\begin{table*}[!ht]
  \centering
  \resizebox{0.98\textwidth}{!}{
    \begin{tabular}{c|cccccccccccccc}

    \midrule
    \textbf{Method} & \textbf{ashtray0} & \textbf{bag0} & \textbf{bottle0} & \textbf{bottle1} & \textbf{bottle3} & \textbf{bowl0} & \textbf{bowl1} & \textbf{bowl2} & \textbf{bowl3} & \textbf{bowl4} & \textbf{bowl5} & \textbf{bucket0} & \textbf{bucket1} & \textbf{cap0} \\
    \midrule
\textbf{BTF (Raw) (CVPR23’)} & 0.512 & 0.430 & 0.551 & 0.491 & 0.720 & 0.524 & 0.464 & 0.426 & 0.685 & 0.563 & 0.517 & 0.617 & 0.686 & 0.524 \\
\textbf{BTF (FPFH) (CVPR23’)} & 0.624 & 0.746 & 0.641 & 0.549 & 0.622 & 0.710 & 0.768 & 0.518 & 0.590 & 0.679 & 0.699 & 0.401 & 0.633 & 0.730 \\
\textbf{M3DM (CVPR23’)} & 0.577 & 0.637 & 0.663 & 0.637 & 0.532 & 0.658 & 0.663 & 0.694 & 0.657 & 0.624 & 0.489 & 0.698 & 0.699 & 0.531 \\
\textbf{PatchCore (FPFH) (CVPR22’)} & 0.597 & 0.574 & 0.654 & 0.687 & 0.512 & 0.524 & 0.531 & 0.625 & 0.327 & 0.720 & 0.358 & 0.459 & 0.571 & 0.472 \\
\textbf{PatchCore (PointMAE) (CVPR22’)} & 0.495 & 0.674 & 0.553 & 0.606 & 0.653 & 0.527 & 0.524 & 0.515 & 0.581 & 0.501 & 0.562 & 0.586 & 0.574 & 0.544 \\
\textbf{CPMF (PR24’)} & 0.615 & 0.655 & 0.521 & 0.571 & 0.435 & 0.745 & 0.488 & 0.635 & 0.641 & 0.683 & 0.684 & 0.486 & 0.601 & 0.601 \\
\textbf{Reg3D-AD (NeurIPS23’)} & 0.698 & 0.715 & 0.886 & 0.696 & 0.525 & 0.775 & 0.615 & 0.593 & 0.654 & 0.800 & 0.691 & 0.619 & 0.752 & 0.632 \\
\textbf{IMRNet (CVPR24’)} & 0.671 & 0.668 & 0.556 & 0.702 & 0.641 & 0.781 & 0.705 & 0.684 & 0.599 & 0.576 & 0.715 & 0.585 & 0.774 & 0.715 \\
\textbf{ISMP (AAAI25')} & 0.865 & 0.734 & 0.722 & 0.869 & 0.740 & 0.762 & 0.702 & 0.706 & 0.851 & 0.753 & 0.733 & 0.545 & 0.683 & 0.672 \\
\textbf{PO3AD (CVPR25')} & \textcolor{blue}{0.962} & {0.949} & {0.912} & 0.844 & 0.88 & {\textcolor{blue}{0.978}} & \textcolor{red}{0.914} & \textcolor{blue}{0.918} & 0.935 & \textcolor{blue}{0.967} & \textcolor{blue}{0.941} & 0.755 & 0.899 & \textcolor{blue}{0.957} \\
\textbf{DUS-Net (MM25')} & 0.612 & 0.628 & 0.749 & 0.822 & 0.641 & 0.769 & 0.735 & 0.617 & 0.574 & 0.812 & 0.744 & 0.738 & 0.754 & 0.701 \\
\textbf{PASDF (ICCV25')} & 0.919 & \textcolor{blue}{0.958} & 0.951 & \textcolor{blue}{0.926} & \textcolor{blue}{0.948} & 0.963 & 0.900 & 0.816 & 0.939 & 0.865 & 0.909 & \textcolor{blue}{0.875} & 0.824 & 0.948 \\
\textbf{CASL (AAAI26’)} & \textcolor{red}{0.974} & \textcolor{red}{0.971} & \textcolor{red}{0.980} & \textcolor{red}{0.980} & \textcolor{red}{0.996} & 0.931 & \textcolor{blue}{0.909} & 0.876 & \textcolor{red}{0.971} & \textcolor{red}{0.971} & \textcolor{red}{0.971} & \textcolor{red}{0.971} & \textcolor{blue}{0.927} & 0.914 \\
\textbf{Ours} & 0.905 & \textcolor{blue}{0.958} & \textcolor{blue}{0.971} & 0.924 & 0.891 & \textcolor{red}{0.985} & \textcolor{blue}{0.909} & \textcolor{red}{0.919} & \textcolor{blue}{0.953} & 0.933 & 0.907 & 0.791 & \textcolor{red}{0.939} & \textcolor{red}{0.962} \\


    \midrule
    \multicolumn{1}{c}{} &       &       &       &       &       &       &       &       &       &       &       &       &       &  \\
    \midrule
    \textbf{Method} & \textbf{cap3} & \textbf{cap4} & \textbf{cap5} & \textbf{cup0} & \textbf{cup1} & \textbf{eraser0} & \textbf{headset0} & \textbf{headset1} & \textbf{helmet0} & \textbf{helmet1} & \textbf{helmet2} & \textbf{helmet3} & \textbf{jar0} & \textbf{micro.} \\
    \midrule
\textbf{BTF (Raw) (CVPR23’)} & 0.687 & 0.469 & 0.373 & 0.632 & 0.561 & 0.637 & 0.578 & 0.475 & 0.504 & 0.449 & 0.605 & 0.700 & 0.423 & 0.583 \\
\textbf{BTF (FPFH) (CVPR23’)} & 0.658 & 0.524 & 0.586 & 0.790 & 0.619 & 0.719 & 0.620 & 0.591 & 0.575 & 0.749 & 0.643 & 0.724 & 0.427 & 0.675 \\
\textbf{M3DM (CVPR23’)} & 0.605 & 0.718 & 0.655 & 0.715 & 0.556 & 0.710 & 0.581 & 0.585 & 0.599 & 0.427 & 0.623 & 0.655 & 0.541 & 0.358 \\
\textbf{PatchCore (FPFH) (CVPR22’)} & 0.653 & 0.595 & 0.795 & 0.655 & 0.596 & 0.810 & 0.583 & 0.464 & 0.548 & 0.489 & 0.455 & 0.737 & 0.478 & 0.488 \\
\textbf{PatchCore (PointMAE) (CVPR22’)} & 0.488 & 0.725 & 0.545 & 0.510 & {0.856} & 0.378 & 0.575 & 0.423 & 0.580 & 0.562 & 0.542 & 0.615 & 0.487 & {0.886} \\
\textbf{CPMF (PR24’)} & 0.551 & 0.553 & 0.551 & 0.497 & 0.509 & 0.689 & 0.620 & 0.458 & 0.555 & 0.542 & 0.515 & 0.520 & 0.611 & 0.545 \\
\textbf{Reg3D-AD (NeurIPS23’)} & 0.718 & 0.815 & 0.467 & 0.685 & 0.698 & 0.755 & 0.580 & 0.626 & 0.600 & 0.624 & 0.825 & 0.620 & 0.599 & 0.599 \\
\textbf{IMRNet (CVPR24’)} & 0.706 & 0.753 & 0.742 & 0.643 & 0.688 & 0.548 & 0.705 & 0.476 & 0.598 & 0.604 & 0.644 & 0.663 & 0.765 & 0.742 \\
\textbf{ISMP (AAAI25')} & 0.775 & 0.661 & 0.770 & 0.552 & 0.851 & 0.524 & 0.472 & 0.843 & 0.615 & 0.603 & 0.568 & 0.522 & 0.661 & 0.600 \\
\textbf{PO3AD (CVPR25')} & \textcolor{red}{0.948} & \textcolor{red}{0.94} & {0.864} & 0.909 & \textcolor{red}{0.932} & \textcolor{red}{0.974} & {0.823} & {\textcolor{blue}{0.907}} & {0.878} & \textcolor{red}{0.948} & \textcolor{blue}{0.932} & {0.846} & {0.871} & 0.81 \\
\textbf{DUS-Net (MM25')} & 0.763 & 0.783 & 0.844 & 0.727 & 0.654 & 0.569 & 0.718 & 0.749 & 0.718 & 0.737 & 0.744 & 0.682 & 0.771 & 0.648 \\
\textbf{PASDF (ICCV25')} & 0.861 & \textcolor{blue}{0.894} & \textcolor{blue}{0.920} & 0.948 & \textcolor{blue}{0.884} & \textcolor{blue}{0.945} & 0.863 & 0.891 & 0.816 & 0.646 & 0.809 & \textcolor{blue}{0.958} & \textcolor{blue}{0.959} & \textcolor{blue}{0.951} \\
\textbf{CASL (AAAI26’)} & \textcolor{blue}{0.894} & 0.853 & 0.602 & \textcolor{blue}{0.958} & 0.796 & 0.732 & \textcolor{red}{0.972} & 0.886 & \textcolor{blue}{0.887} & 0.862 & \textcolor{red}{0.965} & 0.821 & 0.838 & 0.797 \\
\textbf{Ours} & {0.873} & 0.891 & \textcolor{red}{0.957} & \textcolor{red}{0.971} & 0.841 & {\textcolor{blue}{0.945}} & \textcolor{blue}{0.940} & \textcolor{red}{0.974} & \textcolor{red}{0.913} & \textcolor{blue}{0.944} & 0.859 & \textcolor{red}{0.971} & \textcolor{red}{0.977} & \textcolor{red}{0.959} \\

    \midrule
    \multicolumn{1}{c}{} &       &       &       &       &       &       &       &       &       &       &       &       &       &  \\
    \midrule
    \textbf{Method} & \textbf{shelf0} & \textbf{tap0} & \textbf{tap1} & \textbf{vase0} & \textbf{vase1} & \textbf{vase2} & \textbf{vase3} & \textbf{vase4} & \textbf{vase5} & \textbf{vase7} & \textbf{vase8} & \textbf{vase9} & \textbf{Average}\\
    \midrule
\textbf{BTF (Raw)  (CVPR23’)} & 0.464 & 0.527 & 0.564 & 0.618 & 0.549 & 0.403 & 0.602 & 0.613 & 0.585 & 0.578 & 0.550 & 0.564 & 0.550 \\
\textbf{BTF (FPFH) (CVPR23’)} & 0.619 & 0.568 & 0.596 & 0.642 & 0.619 & 0.646 & 0.699 & 0.710 & 0.429 & 0.540 & 0.662 & 0.568 & 0.628 \\
\textbf{M3DM (CVPR23’)} & 0.554 & 0.654 & 0.712 & 0.608 & 0.602 & 0.737 & 0.658 & 0.655 & 0.642 & 0.517 & 0.551 & 0.663 & 0.616 \\
\textbf{PatchCore (FPFH)  (CVPR22’)} & 0.613 & 0.733 & {0.768} & 0.655 & 0.453 & 0.721 & 0.430 & 0.505 & 0.447 & 0.693 & 0.575 & 0.663 & 0.580 \\
\textbf{PatchCore (PointMAE) (CVPR22’)} & 0.543 & {0.858} & 0.541 & 0.677 & 0.551 & 0.742 & 0.465 & 0.523 & 0.572 & 0.651 & 0.364 & 0.423 & 0.577 \\
\textbf{CPMF (PR24’)} & 0.783 & 0.458 & 0.657 & 0.458 & 0.486 & 0.582 & 0.582 & 0.514 & 0.651 & 0.504 & 0.529 & 0.545 & 0.573 \\
\textbf{Reg3D-AD (NeurIPS23’)} & 0.688 & 0.589 & 0.741 & 0.548 & 0.602 & 0.405 & 0.511 & 0.755 & 0.624 & 0.881 & 0.811 & 0.694 & 0.668 \\
\textbf{IMRNet (CVPR24’)} & 0.605 & 0.681 & 0.699 & 0.535 & 0.685 & 0.614 & 0.401 & 0.524 & 0.682 & 0.593 & 0.635 & 0.691 & 0.650 \\
\textbf{ISMP (AAAI25')} & 0.701 & 0.844 & 0.678 & 0.687 & 0.534 & 0.773 & 0.622 & 0.546 & 0.580 & 0.747 & 0.736 & 0.823 & 0.691 \\
\textbf{PO3AD (CVPR25')} & 0.663 & 0.783 & 0.692 & \textcolor{red}{0.955} & \textcolor{blue}{0.882} & \textcolor{blue}{0.978} & 0.884 & \textcolor{blue}{0.902} & \textcolor{red}{0.937} & \textcolor{red}{0.982} & \textcolor{blue}{0.950} & \textcolor{blue}{0.952} & 0.898 \\
\textbf{DUS-Net (MM25')} & 0.740 & 0.728 & 0.743 & 0.699 & 0.648 & 0.650 & 0.731 & 0.765 & 0.711 & 0.728 & 0.762 & 0.581 & 0.712 \\
\textbf{PASDF (ICCV25')} & \textcolor{blue}{0.865} & 0.884 & 0.902 & \textcolor{blue}{0.944} & 0.797 & 0.956 & 0.868 & 0.899 & 0.915 & 0.959 & 0.909 & 0.863 & 0.897 \\
\textbf{CASL (AAAI26’)} & \textcolor{red}{0.997} & \textcolor{blue}{0.929} & \textcolor{red}{0.969} & 0.928 & \textcolor{red}{0.994} & \textcolor{red}{0.988} & \textcolor{red}{0.982} & \textcolor{red}{0.975} & 0.770 & \textcolor{blue}{0.968} & \textcolor{red}{0.982} & \textcolor{red}{0.974} & \textcolor{blue}{0.899} \\
\textbf{Ours} & 0.852 & \textcolor{red}{0.947} & \textcolor{blue}{0.958} & {0.936} & 0.841 & {0.973} & \textcolor{blue}{0.912} & {0.891} & \textcolor{blue}{0.931} & 0.962 & {0.930} & {0.869} & \textcolor{red}{0.924} \\

    \bottomrule
    \end{tabular}%
}
      \caption{The P-AUROC ($\uparrow$) performance of different methods on Anomaly-ShapeNet,where the best and second-place results are highlighted in \textcolor{red}{red} and \textcolor{blue}{blue}, respectively}
  \label{results2}
\end{table*}

\section{Experiments}

\subsection{Experimental setup}
\textbf{Datasets.}
Extensive experiments were conducted on two 3D point cloud datasets, namely Anomaly-ShapeNet~\cite{IMRNet} and Real3D-AD~\cite{Real3D-AD}. Anomaly-ShapeNet contains over 1,600 positive and negative samples across 40 categories, each of which includes only four normal samples for training. Real3D-AD consists of 1,254 large-scale high-resolution point cloud samples from 12 categories, with the training set containing only 4 normal samples per category.

\textbf{Baselines.} Recent methods, including BTF~\cite{BTF}, PatchCore~\cite{Patchcore}, M3DM~\cite{M3DM}, CPMF~\cite{CPMF}, Reg3D-AD~\cite{Real3D-AD}, IMRNet~\cite{IMRNet}, R3D-AD~\cite{R3DAD}, ISMP~\cite{ISMP}, PO3AD~\cite{PO3AD}, DUS-Net~\cite{DU-Net}, PASDF~\cite{PASDF} and CASL~\cite{Casl} were chosen for comparison. The experimental results of these methods were obtained either by running publicly available codes or by fetching from their original papers.

\textbf{Evaluation Metrics.} To evaluate the performance of different anomaly detection methods, the Area Under the Receiver Operator Curve (AUROC, $\uparrow$) was employed to evaluate both object-level anomaly detection and pixel-level anomaly localization performance.  

\textbf{Implementation Details.} Training and evaluation were conducted on a single RTX 3090 GPU (24 GB). In the multi-scale feature extraction module MLF, the base number of divisions in each scale was set to $base\_{lod}=2$, while the total level of LODs was set to $L=5$. The entire model was trained using the Adam optimizer with a learning rate of $1e-4$. Note that all experiments were run independently three times, and their average performance was reported to ensure reliability. 

\subsection{Experimental Results}
\textbf{Results on Anomaly-ShapeNet.} Tables \ref{results1} and \ref{results2} summarize the detection and localization results of different methods on Anomaly-ShapeNet, respectively. Our method achieved an average O-AUROC of 0.921 and an average P-AUROC of 0.924, outperforming the second-best method by 2.1\% and 2.5\%, respectively. These results highlight the robust anomaly detection ability of our method across different categories.

\begin{table*}[!ht]
  \centering
  \resizebox{1\textwidth}{!}{
    \begin{tabular}{c|cccccccccccccc}
    \toprule
    \textbf{Method} & \textbf{Airplane} & \textbf{Car} & \textbf{Candy} & \textbf{Chicken} & \textbf{Diamond} & \textbf{Duck} & \textbf{Fish} & \textbf{Gemstone} & \textbf{Seahorse} & \textbf{Shell} & \textbf{Starfish} & \textbf{Toffees} & \textbf{Average}  \\
    \midrule
\textbf{BTF (Raw) (CVPR23')} & 0.730 & 0.560 & 0.539 & 0.789 & 0.707 & 0.691 & 0.602 & 0.686 & 0.596 & 0.396 & 0.530 & 0.703 & 0.635 \\
\textbf{BTF (FPFH) (CVPR23')} & 0.520 & 0.560 & 0.630 & 0.432 & 0.545 & 0.784 & 0.549 & 0.648 & 0.779 & 0.754 & 0.575 & 0.462 & 0.603 \\
\textbf{M3DM (CVPR23')} & 0.434 & 0.541 & 0.552 & 0.683 & 0.602 & 0.433 & 0.540 & 0.644 & 0.495 & 0.694 & 0.551 & 0.450 & 0.552 \\
\textbf{PatchCore (FPFH) (CVPR22')} & \textcolor{red}{0.882} & 0.590 & 0.541 & 0.837 & 0.574 & 0.546 & 0.675 & 0.370 & 0.505 & 0.589 & 0.441 & 0.565 & 0.593 \\
\textbf{PatchCore (PointMAE)  (CVPR22')} & 0.726 & 0.498 & 0.663 & 0.827 & 0.783 & 0.489 & 0.630 & 0.374 & 0.539 & 0.501 & 0.519 & 0.585 & 0.594 \\
\textbf{CPMF (PR24')} & 0.701 & 0.551 & 0.552 & 0.504 & 0.523 & 0.582 & 0.558 & 0.589 & 0.729 & 0.653 & 0.700 & 0.390 & 0.586 \\
\textbf{Reg3D-AD (NeurIPS23')} & 0.716 & 0.697 & 0.685 & {\textcolor{blue}{0.852}} & 0.900 & 0.584 & 0.915 & 0.417 & 0.762 & 0.583 & 0.506 & 0.827 & 0.704 \\
\textbf{IMRNet (CVPR24')} & 0.762 & 0.711 & 0.755 & 0.780 & 0.905 & 0.517 & 0.880 & 0.674 & 0.604 & 0.665 & 0.674 & 0.774 & 0.725 \\
\textbf{R3D-AD (ECCV24')} & 0.772 & 0.696 & 0.713 & 0.714 & 0.685 & \textcolor{red}{0.909} & 0.692 & 0.665 & 0.720 & 0.840 & 0.701 & 0.703 & 0.734 \\
\textbf{ISMP (AAAI25')} & \textcolor{blue}{0.858} & 0.731 & 0.852 & 0.714 & {0.948} & 0.712 & 0.945 & 0.468 & 0.729 & 0.623 & 0.660 & 0.842 & 0.767 \\
\textbf{PO3AD (CVPR25')} & 0.804 & 0.654 & 0.785 & 0.686 & 0.801 & 0.820 & 0.859 & {0.693} & 0.756 & 0.800 & \textcolor{blue}{0.758} & 0.771 & 0.766 \\
\textbf{DUS-Net (MM25')} & 0.718 & {0.738} & {\textcolor{blue}{0.856}} & 0.696 & 0.824 & \textcolor{blue}{0.844} & 0.908 & \textcolor{blue}{0.733} & 0.814 & 0.822 & {0.755} & 0.834 & 0.795 \\
\textbf{PASDF (ICCV25')} & 0.628 & \textcolor{red}{0.959} & 0.788 & 0.739 & 0.894 & 0.658 & \textcolor{red}{0.989} & 0.634 & \textcolor{red}{1.000} & \textcolor{blue}{0.850} & 0.617 & 0.866 & 0.802 \\
\textbf{CASL (AAAI26’)} & 0.808 & 0.799 & 0.848 & 0.657 & \textcolor{red}{0.976} & 0.836 & 0.935 & \textcolor{red}{0.769} & 0.643 & 0.791 & \textcolor{red}{0.893} & \textcolor{red}{0.924} & \textcolor{blue}{0.823} \\
\textbf{Ours} & 0.674 & \textcolor{blue}{0.872} & \textcolor{red}{0.976} & \textcolor{red}{0.856} & \textcolor{blue}{0.958} & 0.797 & \textcolor{blue}{0.979} & 0.676 & \textcolor{blue}{0.964} & \textcolor{red}{0.922} & 0.710 & \textcolor{blue}{0.920} & \textcolor{red}{0.859} \\


    \bottomrule
    \end{tabular}%
}
    \caption{The O-AUROC ($\uparrow$) performance of different methods on Real3D-AD, where the best and second-place results are highlighted in \textcolor{red}{red} and \textcolor{blue}{blue}, respectively}
  \label{results3}
\end{table*}

\begin{table*}[!ht]
  \centering
  \resizebox{1\textwidth}{!}{
    \begin{tabular}{c|cccccccccccccc} 
    \midrule
    \textbf{Method} & \textbf{Airplane} & \textbf{Car} & \textbf{Candy} & \textbf{Chicken} & \textbf{Diamond} & \textbf{Duck} & \textbf{Fish} & \textbf{Gemstone} & \textbf{Seahorse} & \textbf{Shell} & \textbf{Starfish} & \textbf{Toffees} & \textbf{Average} \\
    \midrule
\textbf{BTF (Raw) (CVPR23')} & 0.564 & 0.647 & 0.735 & 0.609 & 0.563 & 0.601 & 0.514 & 0.597 & 0.520 & 0.489 & 0.392 & 0.623 & 0.571 \\
\textbf{BTF (FPFH) (CVPR23')} & 0.738 & 0.708 & 0.864 & 0.735 & 0.882 & 0.875 & 0.709 & 0.891 & 0.512 & 0.571 & 0.501 & 0.815 & 0.733 \\
\textbf{M3DM (CVPR23')} & 0.547 & 0.602 & 0.679 & 0.678 & 0.608 & 0.667 & 0.606 & 0.674 & 0.560 & 0.738 & 0.532 & 0.682 & 0.631 \\
\textbf{PatchCore (FPFH) (CVPR22')} & 0.562 & 0.754 & 0.780 & 0.429 & 0.828 & 0.264 & 0.829 & \textcolor{blue}{0.910} & 0.739 & 0.739 & 0.606 & 0.747 & 0.682 \\
\textbf{PatchCore (PointMAE) (CVPR22')} & 0.569 & 0.609 & 0.627 & 0.729 & 0.718 & 0.528 & 0.717 & 0.444 & 0.633 & 0.709 & 0.580 & 0.580 & 0.620 \\
\textbf{Reg3D-AD (NeurIPS23')} & 0.631 & 0.718 & 0.724 & 0.676 & 0.835 & 0.503 & 0.826 & 0.545 & 0.817 & 0.811 & 0.617 & 0.759 & 0.705 \\
\textbf{R3D-AD (ECCV24')} & 0.594 & 0.557 & 0.593 & 0.620 & 0.555 & 0.635 & 0.573 & 0.668 & 0.562 & 0.578 & 0.608 & 0.568 & 0.592 \\
\textbf{ISMP (AAAI25')} & 0.753 & 0.836 & \textcolor{blue}{0.907} & 0.798 & 0.926 & \textcolor{blue}{0.876} & 0.886 & 0.857 & 0.813 & 0.839 & 0.641 & \textcolor{blue}{0.895} & 0.836 \\
\textbf{DUS-Net (MM25')} & 0.721 & \textcolor{blue}{0.896} & 0.884 & \textcolor{blue}{0.838} & \textcolor{blue}{0.938} & 0.793 & \textcolor{blue}{0.910} & 0.848 & 0.801 & 0.872 & 0.799 & 0.861 & 0.847 \\
\textbf{PASDF (ICCV25')} & 0.777 & 0.802 & 0.546 & 0.768 & 0.699 & 0.782 & 0.837 & 0.654 & \textcolor{red}{0.887} & 0.648 & 0.703 & 0.838 & 0.745 \\
\textbf{CASL (AAAI26’)} & \textcolor{blue}{0.842} & \textcolor{red}{0.905} & \textcolor{red}{0.932} & 0.713 & \textcolor{red}{0.988} & \textcolor{red}{0.895} & \textcolor{red}{0.935} & \textcolor{red}{0.916} & 0.814 & \textcolor{blue}{0.873} & \textcolor{blue}{0.839} & \textcolor{red}{0.937} & \textcolor{red}{0.882} \\
\textbf{Ours} & \textcolor{red}{0.851} & 0.833 & 0.843 & \textcolor{red}{0.854} & 0.852 & 0.828 & 0.862 & 0.781 & \textcolor{blue}{0.879} & \textcolor{red}{0.893} & \textcolor{red}{0.891} & 0.862 & \textcolor{blue}{0.852} \\


    \bottomrule
    \end{tabular}%
}
    \caption{The P-AUROC ($\uparrow$) performance of different methods on Real3D-AD,where the best and second-place results are highlighted in \textcolor{red}{red} and \textcolor{blue}{blue}, respectively}
  \label{results11}
\end{table*}

\textbf{Results on Real3D-AD.} Tables \ref{results3} and \ref{results11} present the performance of different methods on Real3D-AD. Our method attained an average O-AUROC and P-AUROC performance of 0.859 and 0.852, respectively. Although the pixel-level localization performance was lower than that of the recent CASL method, its object-level detection performance outperformed the CASL method by 3.6\%. These results validate the competitiveness of our method for high-precision point clouds.

%
Unlike existing approaches that rely on discrete group-based or point-based representations, our method learns a continuous surface representation of the point cloud via a discriminative SDF. This representation not only captures global geometric structures but also fully reflects the local topological relationships within point clouds, thereby achieving superior anomaly localization and detection performance.

\subsection{Ablation Studies}
To verify the effectiveness of the core components in our proposed method, we conducted ablation studies on the Real3D-AD dataset, with results presented in Table~\ref{results4}.

 \begin{table}[htb]
  \centering
    \resizebox{0.8\linewidth}{!}{
    \begin{tabular}{c|cccc}
    \toprule
    \textbf{Method} & $M_1$ & $M_2$ & $M_3$ & $M_4$ \\
    \midrule
    NPG & {\texttimes} & {\checkmark} & \checkmark & \checkmark   \\
    MLF   & \checkmark  & \texttimes & \checkmark & \checkmark \\
    ISD   & \checkmark & \checkmark & \texttimes  & \checkmark  \\
    \midrule
    O-AUROC & 0.581 & 0.758 & 0.732 & 0.859 \\
    P-AUROC & 0.513 & 0.676 & 0.735 & 0.852 \\
    \bottomrule
    \end{tabular}%
}
 \caption{Ablation results on the Real3D-AD dataset.}
  \label{results4}
\end{table}

After removing the NPG module, the model trained solely on raw normal point clouds (denoted as $M_1$) exhibited significant performance degradation, with O-AUROC and P-AUROC decreasing by 27.8\% and 33.9\%, respectively. This decline may be attributed to the absence of synthetically generated noise points during training, which limits the model from effectively learning discriminative features that distinguish between normal and anomalous points.
Consequently, the model's sensitivity to anomalies is substantially reduced.

Replacing the MLF module with a feature extractor MLP (denoted as $M_2$) resulted in a performance reduction of 10.1\% in O-AUROC and 17.6\% in P-AUROC, respectively. The reason for this drop may be that a single-scale feature extractor has limitations in capturing global and local geometric structure information of point clouds, while the proposed MLF module can extract multi-scale features to improve anomaly detection performance.

Training SDF-Net using only point coordinates (XYZ) without the fused multi-scale features (denoted as $M_3$) led to a performance reduction of 12.7\% in O-AUROC and 11.7\% in P-AUROC, respectively. A possible explanation is that utilizing informative global and local features is crucial for learning an accurate surface representation, while the proposed ISD module leverages rich multi-scale features to learn surface representation and thus can effectively distinguish between normal and abnormal points.

In summary, each module improves the model performance, and their synergy achieves (denoted as $M_4$) optimal results, validating their necessity for anomaly detection.

\subsection{Parameter Sensitivity Analysis}
To examine the parameter effects on performance, sensitivity experiments were conducted on the Real3D dataset.

\textbf{Effects of sampling ratio $\mathbf{\alpha}$.} In the NPG module, the sampling ratio of surface, near-surface, and uniform points is determined by a set of parameters $\mathbf{\alpha} = (\alpha_\text{surf}, \alpha_\text{near}, \alpha_\text{uni})$. In the experiments, the base number of surface points was set to 100,000. The results of our method on Real3D when varying the parameters are presented in Table \ref{NoiseParameter}. 

As shown in the table, the parameter $\mathbf{\alpha}$ has a considerable impact on model performance. Increasing the proportion of near-surface points improves model detection performance, whereas a higher ratio of uniform points leads to significant performance degradation. This is because near-surface points facilitate the learning of a discriminative SDF-Net, whereas an excessive number of uniform points interferes with the model's ability to capture key geometric features. Therefore, an appropriate configuration of this ratio is essential for learning a discriminative SDF network to effectively separates normal and abnormal points. In the experiments, these parameters were set to $\mathbf{\alpha} = (2, 2, 1)$, which yielded the best performance. Additional experimental results are provided in the supplementary material.

 \begin{table}[!ht]
  \centering
  \resizebox{0.95\columnwidth}{!}{
    \begin{tabular}{c|ccccc}
    \toprule
    \textbf{$\alpha$} & 2:1:1 & 2:1:2 & 2:2:1 & 2:2:2 & 2:2:3  \\
    \midrule
    O-AUROC & 0.815&0.834&0.859& 0.827& 0.806\\
    P-AUROC & 0.834&0.847&0.852& 0.821& 0.819 \\
    \bottomrule
    \end{tabular}%
 }
   \caption{Effects of the sampling ratio $\mathbf{\alpha} = (\alpha_\text{surf}, \alpha_\text{near}, \alpha_\text{uni})$ on performance. }
  \label{NoiseParameter}
\end{table}
\textbf{Effects of the number of feature levels $L$.} 
In the MLF module, the parameter $L$ controls the number of feature levels. The performance of our method with different numbers of $L$ is reported in Table \ref{LODParameter}. It is evident that increasing the value of $L$ from 1 to 5 improves O-AUROC by 1.4\% and P-AUROC by 1.6\%, respectively, as more feature levels provide richer multi-scale features. However, excessive levels significantly increase the number of model parameters, leading to higher memory consumption and slower inference speed. To achieve an optimal trade-off between performance and efficiency, the number of feature levels was fixed to $L=5$ in all experiments.

 \begin{table}[htb]
  \centering
  \resizebox{0.95\columnwidth}{!}{
    \begin{tabular}{c|ccccc}
    \toprule
    \textbf{LOD ($L$)} & 1&2 & 3 &4  &5\\
    \midrule
    O-AUROC & 0.845&0.854 & 0.853&0.857&0.859\\
    P-AUROC & 0.836&0.841 &0.848&0.851&0.852\\
    \bottomrule
    \end{tabular}%
 }
   \caption{Effects of the number of levels $L$ on performance. }
  \label{LODParameter}
\end{table}
\vspace{-10pt}

\subsection{Robustness to Noisy Data}
\begin{figure}[!htb]
  \centering
  \includegraphics[width=\columnwidth]{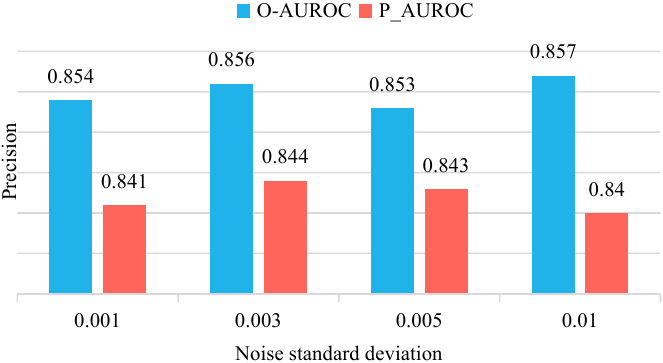}
  \caption{Robustness of the proposed method to test noise.}
  \label{Noise standard deviation}
\end{figure}

In practical application scenarios, the complexity of the environment and the instability of devices often lead to noise interference in scanned point clouds. To evaluate the robustness of the proposed method to noisy data, experiments were conducted on the Real3D dataset, where Gaussian noise with a standard deviation of 0, 0.001, 0.003, 0.005, or 0.01 was injected into test samples (a standard deviation of 0 indicates noise-free data). The experimental results are in Figure~\ref {Noise standard deviation}.

Notably, as the noise standard deviation increases, the model performance degrades moderately. Moreover, even in its worst-case noise scenario, the proposed method still outperforms other comparative methods that use clean data without injecting noise. These empirical findings highlight the robustness of our approach to noisy data.

\subsection{Qualitative Results}
\begin{figure}[htb]
  \centering
  \includegraphics[width=\columnwidth]{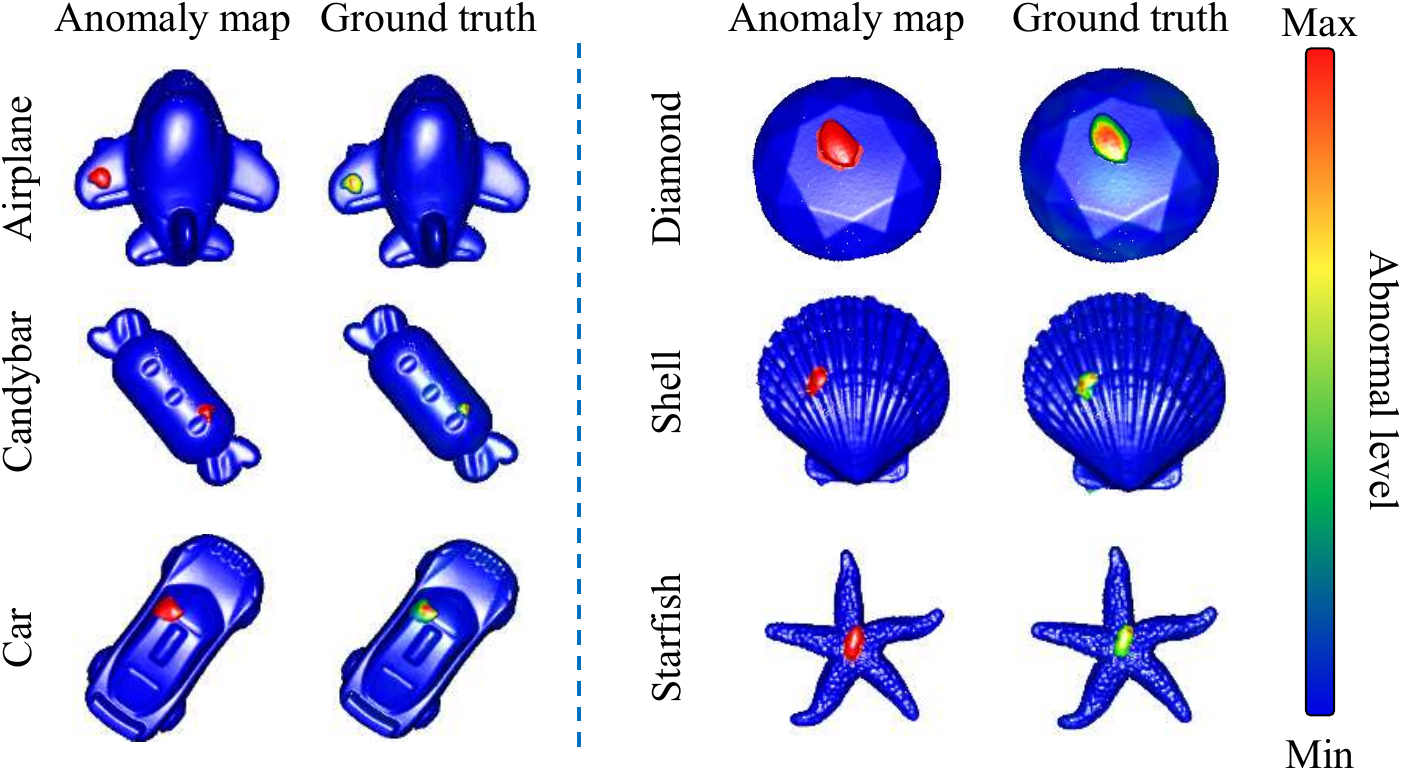}
  \caption{Visualization of detected spatial anomalies in Real3D-AD.}
  \label{vis}
\end{figure}

Figure~\ref{vis} illustrates the anomaly localization results across six categories from the Real3D-AD dataset, where deeper red colors indicate a higher probability of being anomalies. It is observed that our method accurately detects and localizes anomalous regions within point clouds, showcasing its effectiveness in 3D anomaly localization.

\section{Conclusion}
This study presents a surface-based framework for 3D point cloud anomaly detection. By leveraging signed distance function-based surface representation, the proposed method addresses the limitations of previous methods that struggle to integrate global context with local details. The introduced Noisy Points Generation (NPG) module generates both surface points and noisy points, which not only alleviates point cloud sparsity and non-uniformity but also enables the extraction of discriminative features. The Multi-scale Level-of-detail Feature (MLF) module effectively captures hierarchical global and local information to facilitate the training of the anomaly detection model. Furthermore, the Implicit Surface Discrimination (ISD) module learns an accurate surface representation of the point cloud by exploiting multi-scale features, enabling it to distinguish between normal and abnormal points. Extensive experiments demonstrate the superiority of our method, achieving state-of-the-art performance on two benchmark datasets for 3D anomaly detection. Future work will focus on improving the model’s generalization ability and exploring its application in multi-category 3D anomaly detection.
\bibliographystyle{ACM-Reference-Format}
\bibliography{ref} 
\appendix 

\end{document}